\documentclass[runningheads,oneside]{llncs}

\usepackage{eccv}

\usepackage{eccvabbrv}

\usepackage{graphicx}
\usepackage{booktabs}
\usepackage{colortbl} % 支持表格中的颜色
\usepackage{multirow} % 支持单元格合并
\usepackage{xcolor} % 支持颜色定义 
\usepackage{caption} % 支持标题格式
\usepackage{makecell}
\usepackage{listings}
\usepackage[ruled,vlined,linesnumbered]{algorithm2e}
\usepackage{wrapfig}

\newcommand{\lstbg}[3][0pt]{{\fboxsep#1\colorbox{#2}{\strut #3}}}
\lstdefinelanguage{diff}{
  basicstyle=\ttfamily\small,
  morecomment=[f][\lstbg{red!20}]-,
  morecomment=[f][\lstbg{green!20}]+,
}
\lstdefinelanguage{diffpython}{
  language=diff,
  morekeywords={def,if,else,for,while,return,import,from,as,class,with,try,except,break,finally,raise,lambda,and,or,not,in,is,None,True,False},
  morecomment=[l]{\#},
  morestring=[b]",
  morestring=[b]',
}

\usepackage[accsupp]{axessibility}  % Improves PDF readability for those with disabilities.

\usepackage[breaklinks,colorlinks]{hyperref}

\usepackage{orcidlink}

\usepackage[ruled,vlined]{algorithm2e}
\makeatletter
\def\thanks#1{\protected@xdef\@thanks{\@thanks
      \protect\footnotetext{#1}}}
\makeatother
\begin{document}

\title{Starve to Perceive: Taming Lazy Perception in VLMs with Constrained Visual Bandwidth}

\titlerunning{Starve to Perceive}

\author{Yuhuan Wu\inst{1} \and
Cong Wei\inst{2}  \and
Fangzhen Lin\inst{1} \and
Wenhu Chen\inst{2} \and 
Haozhe Wang\inst{1}$^\dagger$ \thanks{Corresponding Author: jasper.whz@outlook.com} 
}

\authorrunning{Y.~Wu et al.}

\institute{
Hong Kong University of Science and Technology \and
University of Waterloo 
}

\maketitle

\begin{abstract}
  
  % \keywords{Multimodal Large Language Models \and Active Visual Reasoning \and Tool Augmented Visual Language Reasoning}
Vision-Language Models (VLMs) deployed as situated agents in high-resolution visual environments require active perception — the ability to dynamically decide where to look through operations like zooming, cropping, and panning. However, current training paradigms produce models that mimic the surface form of such operations without functionally depending on their outputs, a phenomenon we term lazy perception. We trace this to a fundamental learning asymmetry: when coarse global views combined with language priors suffice for moderate accuracy, the model has no incentive to learn harder multi-step visual search. If a model can succeed without actively looking, it will never learn to look. This motivates Starve to Perceive, a training paradigm that constrains visual bandwidth — restricting each observation to a tight token budget so that no single view suffices for task completion, thereby strongly incentivizing strategic and efficient active perception. Despite requiring no auxiliary losses, reward shaping, or architectural changes — serving as a minimal, plug-in modification to standard post-training pipelines — models trained under perceptual starvation achieve substantial gains of  5\% average relative improvement across diverse benchmarks. Our codes and data will be publicly available at \url{https://github.com/WhuanY/Starve2Perceive}.
  \end{abstract}
% \section{Introduction}
% \label{sec:intro}

% This document serves as an example submission to ECCV \ECCVyear{}.
% It illustrates the format authors must follow when submitting a paper. 
% At the same time, it gives details on various aspects of paper submission, including preservation of anonymity and how to deal with dual submissions.
% We advise authors to read this document carefully.

% The document is based on Springer LNCS instructions as well as on ECCV policies, as established over the years.
\section{Introduction}

Vision-Language Models (VLMs) are evolving from passive encoders of static images into situated agents that must operate in complex, high-resolution visual environments. Applications such as autonomous navigation~\cite{anderson2018visionandlanguagenavigationinterpretingvisuallygrounded}, GUI manipulation~\cite{wang2021screen2wordsautomaticmobileui}, and fine-grained document analysis~\cite{masry2022chartqabenchmarkquestionanswering, feng2024docpediaunleashingpowerlarge} demand visual understanding at a level of detail that no single-pass encoding can efficiently provide. Just as biological vision relies on saccadic eye movements to sequentially sample high-acuity information from a vast visual field~\cite{ibbotson2011visual}, effective visual agents must possess \emph{active perception}: the capacity to dynamically determine \emph{where} to look and \emph{what} to examine through deliberate operations such as zooming, cropping, and panning. We argue that active perception is not an optional enhancement but a foundational capability for the next generation of vision-language systems.

Despite growing recognition of this need, current training paradigms fail to produce models that genuinely rely on active visual operations. Supervised fine-tuning (SFT) on tool-use trajectories teaches models to mimic the surface form of perception actions — generating syntactically correct zoom or crop commands — without internalizing functional dependence on their outputs. Standard reinforcement learning (RL), while more flexible, does not resolve this deficiency. Prior works provide concrete diagnostic evidence of a pervasive lazy perception phenomenon: (i) models trained with existing methods maintain near-identical accuracy when intermediate visual observations are replaced with blank inputs, revealing that apparent multi-step visual reasoning is largely illusory~\cite{virl}; and (ii) models seamlessly continue generating correct answers after tool calls are deliberately corrupted, indicating reliance on language priors rather than visual evidence~\cite{pixelreasoner}. 

The root cause is a fundamental asymmetry in learning difficulty. Neural networks follow the path of steepest descent toward loss reduction~\cite{rumelhart1986learning}. When a coarse global view combined with strong language priors suffices to achieve moderate accuracy, the model has no gradient-based reason to invest in the harder strategy of multi-step visual search. Active perception, under standard training, is a \emph{dominated strategy}: it incurs additional computation, introduces the risk of selecting uninformative regions, and yields marginal reward improvements that are overwhelmed by the simpler passive baseline. The critical implication is that \textbf{if the model can succeed without actively looking, it will never learn to look}. Conversely, this asymmetry points directly to a remedy: the passive shortcut must be made structurally unavailable during training.

This analysis motivates our central thesis: \emph{to learn active perception, a model must be trained under conditions where passive observation is insufficient}. We propose \textbf{Starve to Perceive}, a training paradigm grounded in a simple mechanism we term \emph{constrained visual bandwidth}: at each reasoning step, the visual observation returned to the model is rescaled to fit within a tight token budget, ensuring that the information available from any single view is too impoverished for reliable task completion. Under this regime, the model faces a stark choice: either learn to strategically direct its limited visual budget toward task-relevant regions through deliberate zoom, crop, and pan operations, or fail. Active perception ceases to be one strategy among many and becomes the \emph{only viable} strategy. The mechanism requires no auxiliary losses, reward shaping, or architectural modifications, and integrates directly into standard SFT-then-RL post-training pipelines as a reusable, plug-in component.

Crucially, we find that models trained under this perceptual starvation regime acquire active visual reasoning skills that robustly transfer to unconstrained test-time settings: under standard evaluation, our model surpasses the strongest tool-augmented baseline with 3\% higher overall score than the strongest baseline at the same scale. Under test-time resource constraints, our model degrades only marginally due to robust active visual operations, whereas rival baselines suffer drops much larger. Visual bandwidth restriction yields a natural efficiency dividend: fewer visual tokens per step reduce the computational cost of both forward passes and gradient computation, cutting total training time by 50\%, and opens up a path toward deploying active visual agents in compute-constrained settings, including on-device applications and real-time systems~\cite{wang2025bivlmpushingultralowprecision}.

\textbf{Contributions.} We identify visual bandwidth restriction as a principled and effective mechanism for learning functional reliance in VLMs, and instantiate this insight as \emph{Starve to Perceive} — a mechanism that integrates into standard SFT-then-RL pipelines with minimal modifications. We provide the first evidence that active perception skills learned under constrained training transfer to unconstrained evaluation, achieving state-of-the-art accuracy while remaining robust under strict token budgets — a property that promotes functionally grounded visual reasoning and mitigates superficial tool-use mimicry.

\begin{figure*}[t]
    \centering
    \includegraphics[width=\textwidth]{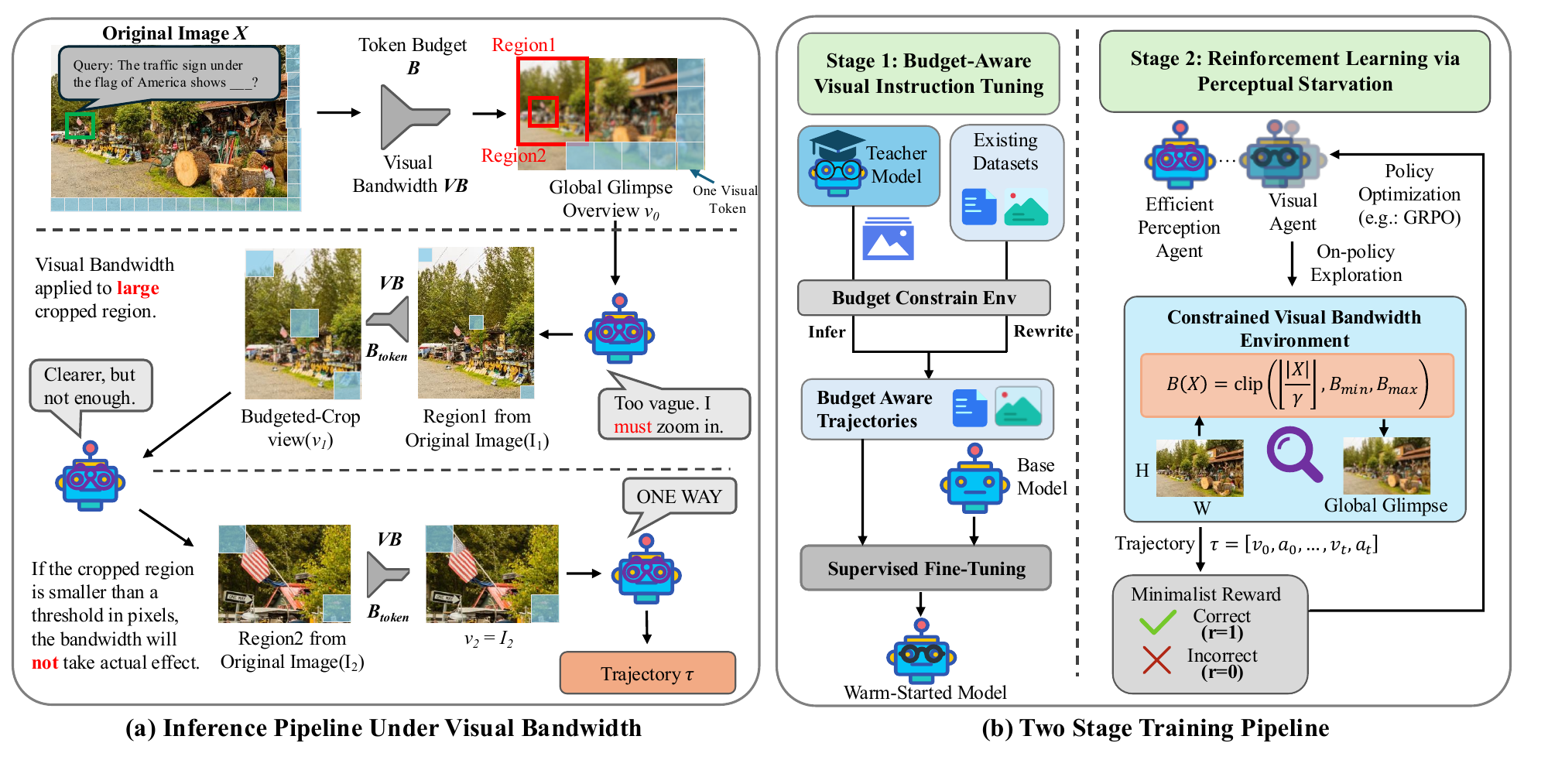}
    \caption{Overview of \textbf{Starve to Perceive}. (a) A Visual Bandwidth (parametrized by $B$) limits the upper bound of both the global image and cropped regions (b) Two-stage training: Budget-Aware Visual Instruction Tuning initializes exploration under token constraints; Reinforcement Learning with Perceptual Starvation train the model via self-collected trajectories under visual constrain to learn active perception.}
    \label{fig:visual_bandwidth_overview}
\end{figure*}

\section{Method}
Vision-Language Models typically process visual information in a single, unconstrained pass. When the entirety of a high-resolution image is available, the path of steepest descent during optimization naturally favors exploiting coarse global context and strong language priors. To overcome this ``lazy perception'' phenomenon, we propose a framework that fundamentally alters the model's observation space. By strictly limiting the visual bandwidth available at any single interaction step, we shift active perception from an optional enhancement to an absolute prerequisite for task success. The overall framework can be seen at Fig~\ref{fig:visual_bandwidth_overview}.
\subsection{The Principle of Perceptual Starvation}

Current models are trained by drinking from a ``firehose'' of high-resolution image data. We propose forcing the model to perceive the environment through a ``straw'' -- a sequence of strictly low-resolution, low-bandwidth glimpses.

This design provides a profound optimization pressure that mimics the Information Bottleneck principle~\cite{tishby2015deeplearninginformationbottleneck}. In standard settings, we want the model's final reasoning state $Z_T$ to maximize mutual information with the correct answer $Y$, denoted as $I(Z_T; Y)$. However, by restricting the maximum token count per glimpse, we introduce a strict upper bound on the channel capacity between the original high-resolution image $X$ and the model's internal state.

Because the ``pipe'' is intentionally narrowed, the current glimpse $v_t$ can only carry a tiny fraction of the total spatial information. If the policy fills this limited capacity with irrelevant background ``noise,'' the predictive power $I(Z_T; Y)$ mathematically drops to zero. Consequently, the only viable mathematical solution to maximize the objective is to learn a policy that actively filters out noise before it enters the context window. By artificially inducing perceptual starvation, active attention transforms from a ``nice-to-have'' feature into a survival mechanism necessary to solve the task.

\subsection{Training with Constrained Visual Bandwidth}
\label{subsec:inference_pipeline}
In this section, we formalize interaction under constrained visual bandwidth and describe how the model answers image queries by progressively acquiring task-relevant visual evidence.

\textbf{Notations.} Let an input be $(X, Q)$, where $X$ is an image and $Q$ is a text query. We denote $\mathcal{D}(\cdot, B)$ as a deterministic downsampling operator that maps any view to a representation whose visual-token count does not exceed a preset budget $B$. Let $v_t$ be the returned visual observation at step $t$, $a_t$ the spatial action (e.g., a bounding box relevant to the image), $Z_t$ the model's internal state, and $H_t=\{Q,v_0,a_1,v_1,\dots,a_{t-1},v_{t-1}\}$ the interaction history.

\textbf{Multi-Turn Interaction.} Inference starts from a global, budgeted overview: $v_0 = \mathcal{D}(X, B)$. At each step $t$, the policy predicts a action conditioned on the current state and query: $a_t = \pi_\theta(Z_{t-1}, Q)$. When $Z_{t-1}$ contains sufficient information for the policy to infer answer, $a_t$ may be just the $Y$. Otherwise, it represents a crop request.

Crucially, to avoid compounding information loss across turns, cropping is always executed on the original image $X$ in native pixel space, and the budget is enforced only after cropping:
\begin{equation}
v_t = \mathcal{D}(\text{Crop}(X, a_t), B).
\end{equation}
This mechanism guarantees that while the model perceives a much smaller spatial area, it does so at a significantly higher effective resolution, seamlessly trading spatial coverage for fine-grained evidence. This constrained observation sequence constitutes the training environment that both SFT (§2.3) and RL (§2.4) operate within.

\subsection{Budget-Aware Visual Instruction Tuning}

Base VLMs lack the inherent ability to emit syntactically correct spatial actions or reason over degraded, bandwidth-constrained images. Because standard supervised fine-tuning (SFT) utilizes full-resolution images, it introduces a severe distribution shift. We bootstrap the policy using a Budget-Aware visual instruction tuning stage, sourcing trajectories for fine-tuning via two complementary methods:

\begin{itemize}
    \item \textbf{Teacher Model Distillation:} We deploy a strong teacher model within our constrained environment to generate multi-turn trajectories. We apply rigorous rejection sampling to retain only trajectories that produce correct final answers and exhibit strict token-efficiency.
    \item \textbf{Trajectory Rewriting:} We convert existing high-quality tool-use datasets into budget-aware formats by applying ``Visual Reconstruction'' (downsampling all intermediate views to the budget $B$) and ``Textual Grounding Realignment'' (remapping spatial coordinates in the reasoning text to match the constrained image space). This can be achieved by a rewriting model(i.e., GPT-4o) via explicit prompting and post-hoc rule-based filtering. We will disclose the rewriting details in the supplementary material.
\end{itemize}

Given a collected trajectory $\tau = (v_0, a_1, v_1, \dots, a_T, Y)$ generated under the budget $B$, we optimize the model parameters $\theta$ by minimizing the negative log-likelihood of the actions and the final answer:
\begin{equation}
\mathcal{L}_{SFT}(\theta) = -\mathbb{E}_{\tau}\left[\sum_{t=1}^{T} \log \pi_{\theta}(a_t|H_t) + \log \pi_{\theta}(Y|H_{T+1})\right]
\end{equation}

\subsection{Reinforcement Learning via Perceptual Starvation}

While SFT provides a necessary behavioral prior, it is fundamentally limited by the perception strategies of the teacher. To develop genuine functional dependence on its own visual observations, the model must autonomously explore the constrained environment through reinforcement learning. 

\textbf{Resolution-Conditioned Token Budget:} To ensure the bandwidth constraint acts as a consistent pressure mechanism -- neither trivially easy for small images nor completely destructive for massive ones -- we implement a resolution-conditioned token budget. Rather than fixing a static scalar, the budget $B$ scales dynamically relative to the native token count of the source image $|X|$:
\begin{equation}
B(X) = \text{clip}\left( \left\lfloor \frac{|X|}{\gamma} \right\rfloor, B_{\min}, B_{\max} \right)
\end{equation}

Here, the compression rate $\gamma \gg 1$ ensures a uniform degree of perceptual starvation across diverse inputs, while $B_{\min}$ and $B_{\max}$ bound the operational extremes.

The crucial advantage of our perceptual starvation formulation is the radical simplification of the reward signal. In unconstrained settings, models require dense, hand-crafted auxiliary losses to penalize excessive API calls or reward specific focus behaviors. In our framework, the constrained environment acts as a strict physical regularizer. As shown in Listing~\ref{lst:budgeted_rollout_python}, integration into existing multi-turn VLM RL pipelines is minimal: we only replace the visual input with its budget-constrained counterpart (highlighted in green in the pseudo-code) in the rollout function. No architectural changes or specialized losses are required, making the method plug-and-play and easy to reuse across different training setups.

\begin{lstlisting}[
    caption={Comparison between a typical multi-turn VLM rollout (red) and our implementation (green). \texttt{B} is the token budget constant, which defines the maximum visual token count per image. \texttt{D} is the deterministic operator that apply image transformation. \texttt{pi} refers to the policy model.},
    label={lst:budgeted_rollout_python},
    abovecaptionskip=2pt,
    belowcaptionskip=7pt,
    language=diffpython
]
- def vanilla_rollout(X, Q, T, pi):
+ def budgeted_rollout(X, Q, B, T, pi, D):
-   v0 = X
+   v0 = D(X, B)
    # first image uses a global glimpse under budget B
    Z = initialize_state(v0, Q)
    for t in range(1, T + 1):
        a_t = pi(Z, Q)
        if has_answer(a_t):
            break
        X_crop = crop(X, a_t)
-       v_t = X_crop
+       v_t = D(X_crop, B)
        # intermediate image is also constrained by budget B
        Z = update_state(Z, v_t)
    Y_hat = pi(Z, Q) # Final answer
    return Y_hat
\end{lstlisting}

Since the constrained environment already acts as a regularizer, we employ a minimalist, sparse reward function $R(\tau)$ for any trajectory $\tau$ generated during rollout:
\begin{equation}
R(\tau) = \begin{cases} 1, & \text{if } Y \text{ exactly matches } Y^* \\ 0, & \text{otherwise} \end{cases}
\end{equation}
Because the visual budget strictly forbids resolving fine details from the global glimpse $v_0$ alone, a $+1$ reward can realistically only be achieved if the intermediate actions $a_t$ successfully isolate task-critical regions. 

We define our training target as to maximize the expected return
\begin{equation}
\max_{\theta} J(\theta) = \mathbb{E}_{Q \sim D, \tau \sim \pi_{\theta}}[R(\tau)]
\end{equation}
Under this objective, active multi-step visual reasoning ceases to be an optional strategy; it becomes the singular pathway to maximizing the reward. In practice, we optimize the return with GRPO~\cite{shao2024deepseekmathpushinglimitsmathematical} algorithm with a clipped surrogate objective.

\section{Experiments}
\label{sec:experiments}
In this section, we first outline the training and evaluation settings. We then examine the effectiveness of our model trained under the constrained visual bandwidth. Finally, we share the key insights for incentivizing precise visual perception for Multi-modal Large Language Models.

\subsection{Setup}

\begin{table*}[tbp]
    \small
    \centering
    \caption{Main results on multimodal benchmarks regarding visual search and perception-intensive reasoning. MRWL and RWQA denote MME-RealWorld-Lite and RealWorldQA, respectively. Best and second best results in each group are highlighted in \textbf{bold} and \underline{underlined}, respectively. $^*$~indicates models capable of calling tools. $^\dagger$~denotes results reported from the other publications; all other entries are reproduced under our unified evaluation protocol.}\label{tab:main_results}
    \scalebox{0.75}{
    \begin{tabular}{lccccccccccccc}
    \toprule
    \multirow{2.5}{*}{\textbf{Model}} & \multicolumn{2}{c}{\textbf{HR-Bench}} & \multicolumn{3}{c}{\textbf{V* Bench}} & \multicolumn{3}{c}{\textbf{VisualProbe\textsubscript{test}}} & \multicolumn{3}{c}{\textbf{Perception-Intensive}} & \multirow{2.5}{*}{\textbf{Overall}} \\
    \cmidrule(lr){2-3} \cmidrule(lr){4-6} \cmidrule(lr){7-9} \cmidrule(lr){10-12}
    & \textbf{4K} & \textbf{8K} & \textbf{Attr} & \textbf{Pos} & \textbf{All} & \textbf{Easy} & \textbf{Med} & \textbf{Hard} & \textbf{MRWL} & \textbf{RWQA} & \textbf{TreeBench} & \\
    \midrule
    \multicolumn{13}{c}{\textit{No Constrain Settings}} \\
    GPT-4o$^\dagger$ & 68.0 & 63.9 & - & - & 62.8 & 47.5 & 15.4 & 11.2 & 46.4 & 67.2 & \textbf{46.9} & - \\
    Gemini-2.5-Pro$^\dagger$ & - & - & - & - & 79.2 & - & - & - & - & - & - & - \\
    LLaVA-OneVision$^\dagger$ & 63.0 & 59.8 & 75.7 & 75.0 & 75.4 & - & - & - & 43.7 & 66.3 & - & - \\
    PixelReasoner$^*$ & 71.13 & \underline{69.25} & 85.22 & 80.26 & 83.25 & 58.4 & 29.6 & 28.8 & 47.21 & \textbf{69.02} & 39.0 & 60.10 \\
    DeepEyes$^*$ & 71.50 & 67.88 & 82.61 & \underline{85.53} & 83.77 & 63.83 & \underline{35.07} & 32.07 & 50.81 & \underline{68.89} & 37.5 & 61.77 \\
    ChainOfFocus$^*$ & 71.63 & 67.25 & 86.09 & \textbf{86.84} & \textbf{86.39} & 56.74 & 32.46 & 38.67 & 48.98 & 64.45 & 39.75 & 61.75 \\
    Mini-o3$^{*\ddagger}$ & 57.75 & 47.75 & 72.17 & 80.26 & 75.39 & 19.86 & 25.37 & 19.81 & 32.15 & 52.55 & 30.37 & 46.68 \\
    \midrule
    Qwen2.5-VL$^\dagger$ & 65.25 & 63.0 & 73.9 & 67.1 & 71.2 & 39.0 & 26.0 & 23.9 & 43.0 & 68.5 & 37.0 & 52.53 \\
    \scriptsize + BA-SFT$^*$ & 72.50 & 65.88 & 85.22 & 73.68 & 80.62 & 49.64 & 32.46 & 33.97 & 46.95 & 67.32 & \underline{40.24} & 58.95 \\
    \scriptsize + BA-SFT+VanillaRL$^*$ & \underline{75.25} & 67.75 & \textbf{87.83} & 82.89 & \underline{85.86} & \textbf{69.50} & \textbf{36.57} & \underline{42.45} & \underline{53.78} & 67.19 & \textbf{40.25} & \underline{64.48} \\
    \rowcolor{gray!20}
    Ours$^*$ & \textbf{76.63} & \textbf{70.00} & \underline{86.96} & 84.21 & \underline{85.86} & \underline{68.09} & \underline{34.70} & \textbf{43.40} & \textbf{55.13} & 67.71 & 37.78 & \textbf{64.59} \\
    \midrule
    \midrule
    \multicolumn{13}{c}{\textit{Constrain Setting ($B=256$)}} \\
    PixelReasoner$^*$ & 49.38 & 43.13 & 46.96 & 65.79 & 54.45 & - & - & - & 29.60 & 55.03 & - & - \\
    DeepEyes$^*$ & 57.13 & 51.25 & 60.87 & 64.47 & 62.30 & 37.58 & 13.43 & 15.09 & 40.85 & 63.01 & 35.30 & 45.57 \\
    ChainOfFocus$^*$ & 59.63 & 53.75 & 55.70 & 68.40 & 60.70 & 38.30 & 14.93 & 14.15 & 37.83 & 60.00 & 34.32 & 45.25 \\
    Mini-o3$^{*\ddagger}$ & 51.63 & 40.88 & 53.04 & 64.47 & 57.59 & 17.02 & 17.54 & 0.90 & 27.36 & 48.63 & 25.19 & 36.75 \\
    \midrule
    Qwen2.5-VL & 38.00 & 26.50 & 19.13 & 44.74 & 29.31 & - & - & - & 27.41 & 56.34 & - \\
    \scriptsize + BA-SFT$^*$ & 59.25 & \underline{56.13} & 55.65 & 64.47 & 59.16 & 47.51 & \underline{19.78} & 14.15 & 39.55 & 60.65 & \underline{36.29} & 46.60 \\
    \scriptsize + BA-SFT+VanillaRL$^*$ & \textbf{66.25} & \underline{60.00} & \textbf{66.09} & \underline{69.74} & \textbf{67.54} & \textbf{58.16} & \underline{18.28} & \underline{15.09} & \underline{46.69} & \underline{63.27} & 35.80 & \underline{51.54} \\
    \rowcolor{gray!20}
    Ours$^*$ & \underline{66.13} & \textbf{60.38} & \underline{62.61} & \textbf{72.36} & \underline{66.49} & \textbf{58.16} & \textbf{20.52} & \textbf{19.81} & \textbf{47.37} & \textbf{64.97} & \textbf{37.04} & \textbf{52.35} \\
    \bottomrule
    \end{tabular}}
    \par\raggedright\footnotesize{$^\ddagger$~Mini-o3 exhibits notably lower performance as the model has a strong bias toward producing many reasoning turns before reaching an answer.}
\end{table*}

\paragraph{\textbf{Implementation Details.}}
We chose Qwen2.5-VL-7B ~\cite{bai2025qwen25vltechnicalreport} as our base policy model. For training data, we reuse image-query sources from prior related works~\cite{pixelreasoner, deepeyes, minio3, treebench} and filter them by two criteria: (i)~image in query must have sufficient high-resolution, because large images widen the gap between the constrained view and the original view, placing greater demand on precise localization. (ii)~queries should be of moderate difficulty, so that the reinforcement-learning signal remains informative rather than saturated. For Instruction Tuning phase, our data synthesis pipeline yield 7k budget-aware trajectories. For reinforcement learning, 19k training queries are drawn from the same curated sources after applying the resolution and difficulty filters above. The default hyperparameters we choose for RL training is $\gamma=6.25, B_{min}=169, B_{max}=1337$. We implement a partial version of DAPO~\cite{dapo}. We present more details for dataset filtering criteria, training data composition and hyperparameter choice in our supplementary materials. 

\paragraph{\textbf{Benchmarks.}} 
We classified the selected benchmarks into three families. The first is \emph{ultra-high-resolution understanding}: HR-Bench~\cite{wang2024divideconquercombinetrainingfree} supplies natural images at 4K and 8K resolution that stress a model's capacity to resolve fine-grained details. VisualProbe\textsubscript{test}~\cite{minio3} stratifies questions into Easy, Medium, and Hard splits, enabling us to evaluate model performance at different difficulty levels. The second family is V*~Bench~\cite{wu2023vguidedvisualsearch}, a well-established visual search benchmark with two subtasks: attributes recognition and spatial relation reasoning. We report the subtask accuracies and overall accuracy. The third family evaluates \emph{perception-intensive reasoning}: MME-RealWorld-Lite~\cite{zhang2025mmerealworldmultimodalllmchallenge}, RealWorldQA~\cite{xai2024grok}, and TreeBench~\cite{treebench} mainly composed of real-world images, charts and table images and at non-trivial resolutions with a wide range of question types, posing not only perceptual but also general reasoning challenges to models.
\paragraph{\textbf{Baselines.}}
We primary compare with model families that utilize zoom-in tool-calls: DeepEyes~\cite{deepeyes} is trained via an end-to-end manner. PixelReasoner~\cite{pixelreasoner} and Chain-of-Focus~\cite{zhang2025adaptivechainoffocusreasoningdynamic} first finetune on synthesized data and adopt specialized reward schema to invoke active perception. Mini-o3~\cite{minio3} goes further from these works by meticulously designing an over-turn masking strategy to foster extra-long turn interactions. We also include Qwen2.5-VL~\cite{bai2025qwen25vltechnicalreport}, LLaVA-OneVision~\cite{li2024llavaonevisioneasyvisualtask} and GPT-4o as baselines to evaluate the performance of the model without any tool call.
\paragraph{\textbf{Evaluation Protocol.}} 
We evaluated our model and other baseline models under two environments: an \emph{Unconstrained} setting, where the full-resolution image is available, and a \emph{Constrained} setting, where every observation is subject to a small token budget $B=256$, and the inference pipeline is the same as ~\ref{subsec:inference_pipeline}. In both cases, we use greedy decoding, limit the interaction to at most 3 turns, and set the maximum context length (including visual tokens) at 30k tokens. These restrictions ensure a fair comparison between methods and focus the evaluation not only on accuracy but also on the efficacy of perception rather than exhaustive exploration.

\subsection{Main Results}
\paragraph{\textbf{Results on Unconstrained Settings.}}
As shown in Table~\ref{tab:main_results}, our best model trained under constrain consistently outperforms all 7B-scale competitive baselines across many benchmark categories in the unconstrained setting by a clear margin. On high resolution visual search, our model leads HR-Bench at both resolutions. It achieves the highest VisualProbe Hard score (43.4 vs.\ 38.7 for Chain-of-Focus) and a competitive VisualProbe Easy score (68.1 vs.\ 63.8 for DeepEyes), indicating that visual constrain training substantially sharpens the model's ability to localize and reason under varying levels of perceptual difficulty. On perception-intensive reasoning, our model leads MME-RealWorld-Lite (55.1 vs.\ 50.8 for DeepEyes), further validating broad gains in scene understanding.

A key finding emerges from the train–test discrepancy in evaluation conditions: \emph{perceptual capabilities acquired under the visual bandwidth \textbf{generalize} beyond the constrained regime}. Our best model is trained exclusively under the limited visual bandwidth, yet at inference time all the intermediate images are provided with no such restriction. Despite this mismatch, the model achieves the highest unconstrained average score(64.59) among all 7B models trained with all unconstrained pipeline(other highest: 61.77, DeepEyes). This result indicates that training under perceptual pressure does not confine the model to low-resolution reasoning; rather, the precision demanded by the bandwidth cultivates fine-grained localization skills that transfer effectively to standard, unconstrained evaluation.

\paragraph{\textbf{Results on Constrained Settings.}}
\begin{table}[t]
    \small
    \centering
    \caption{Performance Gain and Loss comparing to the baseline performance of Qwen2.5-VL-7B.}
    \label{tab:graceful_degradation}
    \setlength{\tabcolsep}{5pt}
    \begin{tabular}{lcccc}
    \toprule
    \multirow{2}{*}{\textbf{Model}} & \multicolumn{2}{c}{\textbf{Overall Score}} & $\boldsymbol{\Delta}$\,\textbf{Gain} & $\boldsymbol{\Delta}$\,\textbf{Loss} \\
    \cmidrule(lr){2-3}
     & No-Constrain  & Constrain  & (No Constrain) & (Constrain) \\
    \midrule
    Qwen2.5-VL-7B  & 52.53 & -- & -- & -- \\
    DeepEyes       & 61.77 & 45.57 & +9.24 & -6.96 \\
    Chain-of-Focus & 61.75 & 45.25 & +9.22 & -7.29 \\
    \midrule
    \rowcolor{gray!20}
    \textbf{Ours}  & 64.59 & 52.35 & +12.06 & -0.18 \\
    \bottomrule
    \end{tabular}
\end{table}
Table~\ref{tab:main_results} further reveals a critical asymmetry in how models capable of zooming-in respond when visual information is limited. We contextualize key data points in Table~\ref{tab:graceful_degradation}. Under unconstrained evaluation, all zooming-in methods yield substantial gains over the non-tool baseline Qwen2.5-VL-7B, suggesting that active visual exploration is broadly beneficial. Yet under constrained evaluation, a stark divergence emerges. DeepEyes and Chain-of-Focus, both trained \textbf{without a visual bandwidth constraint} and performs relatively well under no constrain, suffer gains-to-losses reversals of $-6.96$ and $-7.29$ points relative to their untrained base model. In contrast, our model, trained under visual bandwidth constraints, limits this loss to just $-0.18$ points, preserving nearly all of the baseline's predictive power. 

Another key finding emerges from the huge contrast: \emph{optimizing visual perception performance in an unconstrained setting \textbf{generalizes poorly} to a constrained setting.} Models trained with abundant visual tokens learn to use zooming-in as a \emph{supplement} to already-rich perception—a superficial enhancement that collapses once the underlying richness is removed. Our model, by necessity, learns to use zooming-in as a \emph{survival} mechanism: the sole means of recovering fine-grained detail that the budget otherwise forbids.  The results suggest that effective perception strategies must be forged under the very constraints they are meant to overcome.

\subsection{Further Analysis}
\begin{table}[t]
    \centering
    \caption{\textbf{Standard SFT vs.\ Budget-Aware SFT.}
    Both variants are trained on content-equivalent trajectories;
    the sole difference is whether the visual-budget constraint is present.}
    \label{tab:sft-ablation}
    \resizebox{\linewidth}{!}{%
    \begin{tabular}{l cc ccc c c c}
    \toprule
     & \multicolumn{2}{c}{HRBench} 
     & \multicolumn{3}{c}{V*} 
     & \multirow{2}{*}{MRWL}
     & \multirow{2}{*}{RWQA} 
     & \multirow{2}{*}{TreeBench} \\
    \cmidrule(lr){2-3}\cmidrule(lr){4-6}
     & 4K & 8K & dir\_attr & rela\_pos & Overall &  &  &  \\
    \midrule
    Standard SFT
     & 68.75 & 59.88 & 77.39 & 73.68 & 75.92 & 41.06 & 62.09 & 38.52 \\
    BA-SFT
     & \textbf{72.50} & \textbf{65.88} & \textbf{85.22} & 73.68 & \textbf{80.62} & \textbf{46.95} & \textbf{67.32} & \textbf{40.24} \\
    \midrule
     & \textcolor{teal}{+3.75} & \textcolor{teal}{+6.00} & \textcolor{teal}{+7.83} & 0.00 & \textcolor{teal}{+4.70} & \textcolor{teal}{+5.89} & \textcolor{teal}{+5.23} & \textcolor{teal}{+1.72} \\
    \bottomrule
    \end{tabular}}
    \end{table}
Having established that training under visual constraints improves both unconstrained accuracy and constrained robustness, we conduct a staged ablation to isolate the role of each phase: Budget-Aware Visual Instruction Tuning and Reinforcement Learning with Visual Bandwidth. The four findings below form one mechanism chain: budget-aware SFT creates an active-perception prior, this prior is required for RL to avoid collapse, constrained RL then strengthens that prior, and the strengthened behavior finally translates into benchmark gains.

\paragraph{\textbf{Budget-Aware Tuning vs. Standard Tuning.}} 
To determine whether the gains from Budget-Aware Visual Instruction Tuning stem from the specific reasoning trajectories or the \emph{budget-constrained} framing itself, we perform a controlled trajectory ablation. We synthesize a "reversed" version of our budget-aware data: for each trajectory, we restore the full-resolution image and map all cropping coordinates back to the original pixel space (verified via rigorous rule-based checks). This ensures both datasets contain identical visual reasoning content, differing only in the presence of bandwidth constraints.

As shown in Table~\ref{tab:sft-ablation}, when evaluated in a \textbf{non-constrained} setting, the model fine-tuned on budget-constrained data consistently outperforms the one trained on standard (reversed) trajectories across all benchmarks. This suggests that the budget constraint functions as a vital active perception catalyst, forcing the model to internalize the functional dependency between local visual cues and task success. Consequently, we employ the budget-aware checkpoint as the initialization for the RL phase.

\begin{figure*}[t]
    \centering
    \begin{minipage}[t]{0.32\textwidth}
        \centering
        \includegraphics[width=\linewidth,height=0.20\textheight,keepaspectratio]{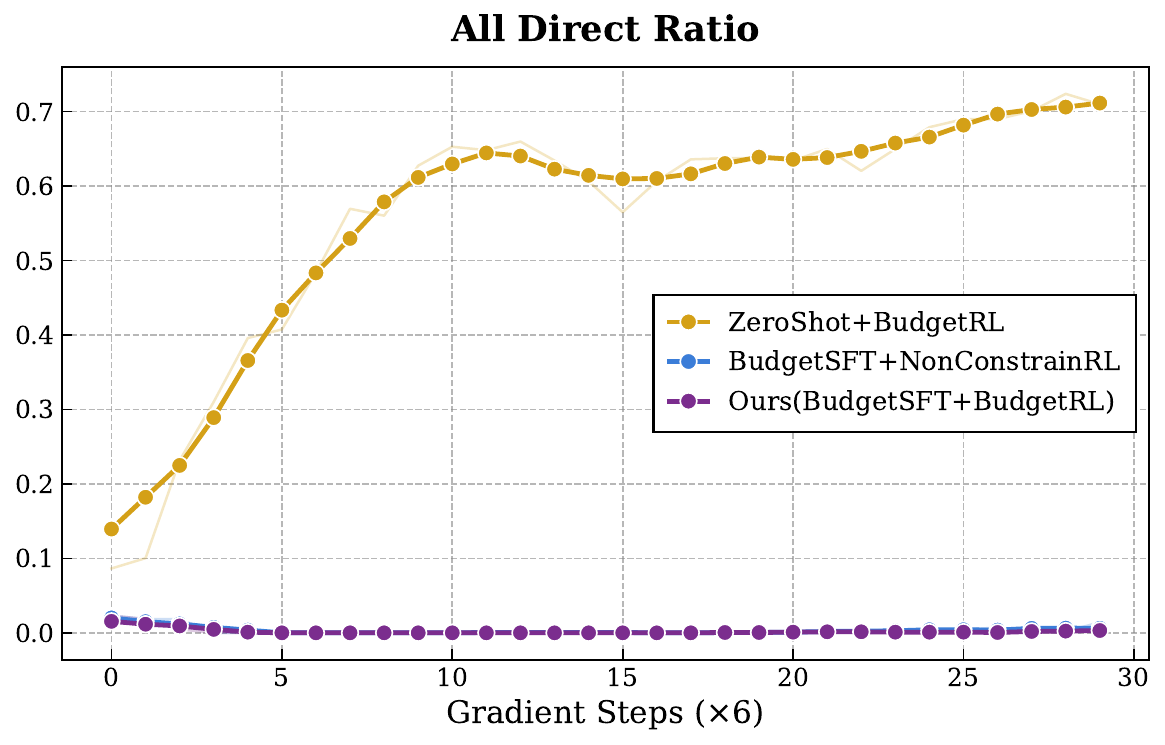}
        \subcaption{All Direct ratio dynamics.}
        \label{fig:direct_ratio}
    \end{minipage}
    \hfill
    \begin{minipage}[t]{0.32\textwidth}
        \centering
        \includegraphics[width=\linewidth,height=0.20\textheight,keepaspectratio]{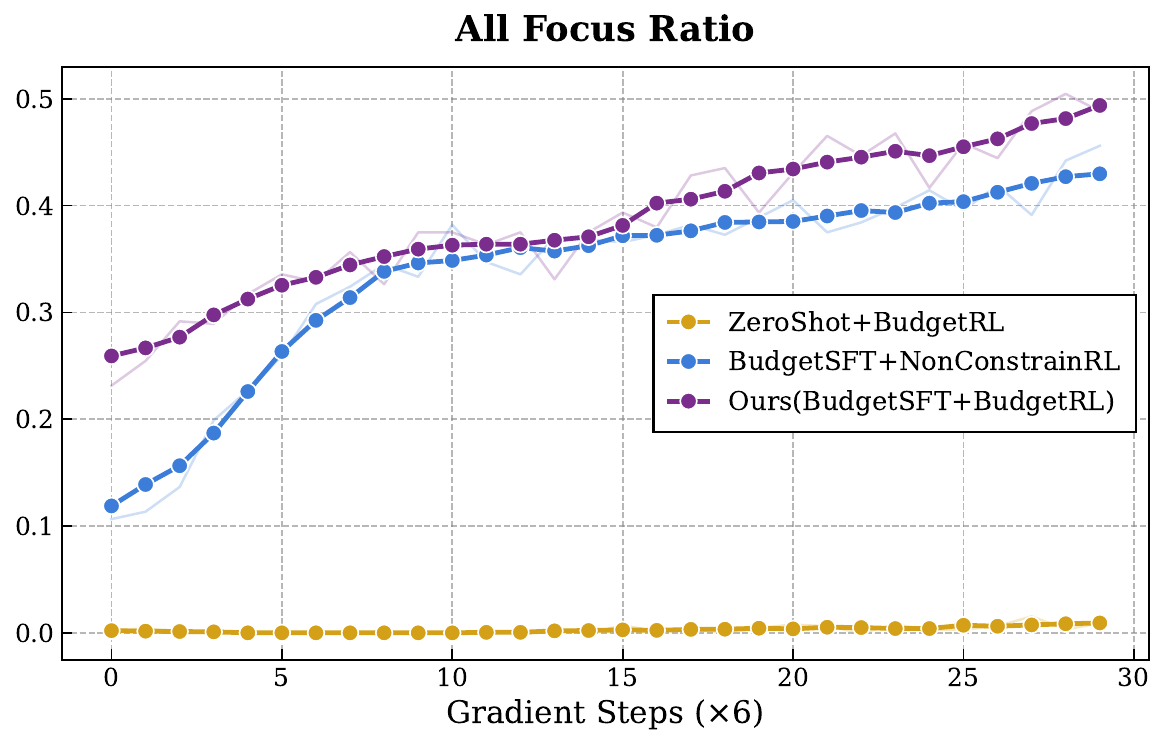}
        \subcaption{All Focus ratio dynamics.}
        \label{fig:focus_ratio}
    \end{minipage}
    \hfill
    \begin{minipage}[t]{0.32\textwidth}
        \centering
        \includegraphics[width=\linewidth,height=0.20\textheight,keepaspectratio]{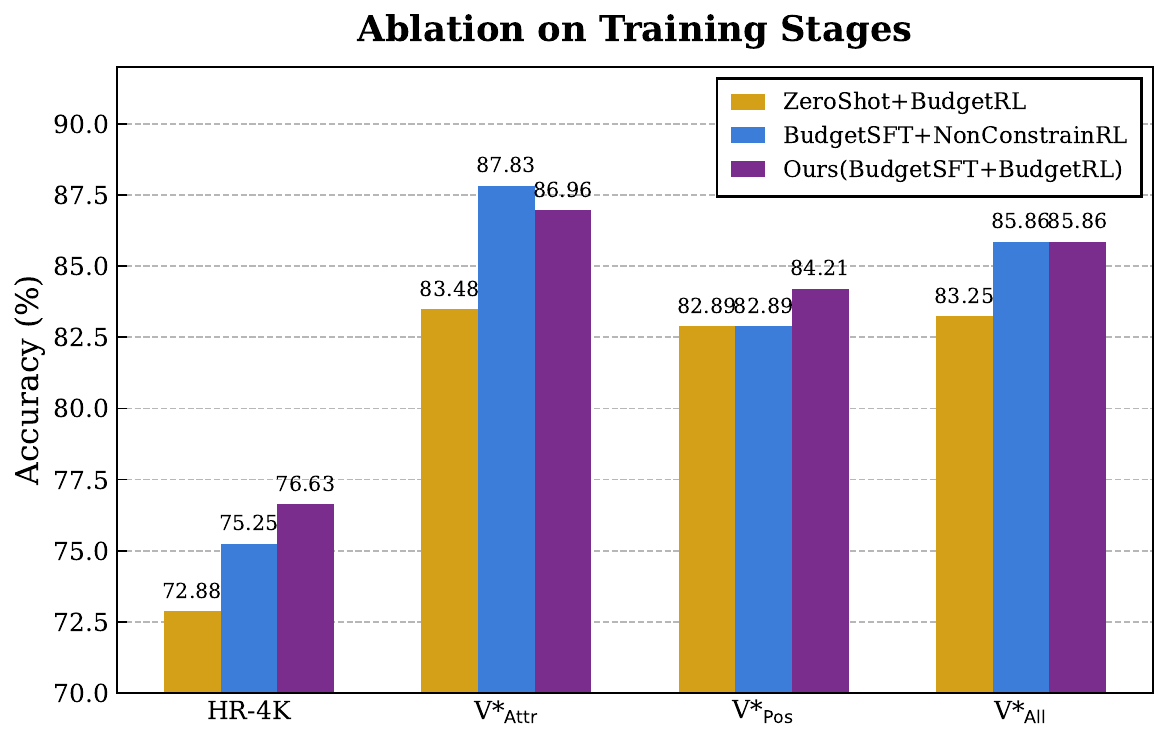}
        \subcaption{Final Performance of Budget Ablation Across Training Stages.}
        \label{fig:ablation_bar}
    \end{minipage}
    \caption{Training Dynamics and Final Performance of Budget Ablation Across Training Stages. "All Direct Ratio" measures the proportion of queries where the model consistently bypasses visual grounding and directly answers across all sampled rollouts at a given policy checkpoint, while "All Focus Ratio" measures the proportion of queries for which the model consistently chooses to select regions across all sampled rollouts.}
    \label{fig:ablation_combined}
\end{figure*}

\paragraph{\textbf{RL Under Constraints Necessitates Visual SFT.}}

We investigate whether a visually constrained environment and accuracy-based rewards are sufficient to elicit active perception from scratch. We compare RL training starting from the base model (\textit{ZeroShot + BudgetRL}) against our SFT-warmed-start model. As illustrated by the yellow lines in Fig.~\ref{fig:direct_ratio} and Fig.~\ref{fig:focus_ratio}, the base model's strategy rapidly collapses: it learns to bypass the costly active perception mechanism, relying instead on language priors to answer questions blindly. By the end of training, active perception is rarely invoked. In contrast, SFT-initialized models (blue and purple lines) maintain and progressively refine their reliance on active perception. This divergence highlights that budget-aware SFT is essential to provide the necessary behavioral prior for successful RL under visual bandwidth constraints.

\paragraph{\textbf{Visual Constraints as an Incentive for Active Perception.}}
To validate whether the visual bandwidth bottleneck effectively incentivizes active perception, we compare RL dynamics in two environments: one without visual constraints (\textit{NonConstrainRL}) and one with them (\textit{BudgetRL}), both starting from the same warm-start checkpoint.

As shown in Fig.~\ref{fig:focus_ratio}, while both models exhibit an upward trend in the $\textit{All Focus Ratio}$, the model trained under visual constraints (purple, "Ours") consistently maintains a significantly higher ratio throughout the training process. By the end of RL, our model achieves an $\textit{All Focus Ratio}$ of approximately 0.50, compared to 0.40 for the unconstrained variant. This indicates that while SFT provides an initial inclination toward active perception, the explicit visual constraint during RL imposes a stronger optimization pressure. This pressure prevents the model from defaulting to "lazy" low-resolution heuristics, instead compelling it to actively ground its reasoning in high-resolution local evidence.

\paragraph{\textbf{Training Dynamics Translate to Downstream Task Performance.}}
The behavioral differences observed in Figures~\ref{fig:direct_ratio} and~\ref{fig:focus_ratio} directly manifest in final benchmark accuracy, as shown in Figure~\ref{fig:ablation_bar}. The model trained without Budget-Aware SFT (\textit{ZeroShot + BudgetRL}), which collapses toward direct answering during RL, achieves the lowest performance across all high-resolution visual search benchmarks. The model trained without the visual constraint during RL (\textit{BA-SFT + VanillaRL}), which exhibits weaker active-perception pressure despite a healthy SFT initialization, yields intermediate scores. Our full pipeline—Budget-Aware SFT followed by RL under the visual bandwidth—achieves the best or competitive accuracy across splits, confirming that the two training stages are complementary rather than redundant: instruction tuning establishes the behavioral foundation for active perception, and the visual constraint during RL amplifies it into measurable task-level gains.

\subsection{Efficiency Analysis.}
\begin{figure*}[t]
    \centering
    \definecolor{DropGreen}{RGB}{34,139,34}
    \definecolor{GainRed}{RGB}{178,34,34}
    %% ---- Left: Table ----
    \begin{minipage}[t]{0.58\textwidth}
        \vspace{0pt}
        \centering
        \captionof{table}{E2E inference latency per sample under identical evaluation infrastructure. \textbf{Vanilla} denotes the base model inferring without token budget. \textbf{Rel.\ Perf.} denotes the relative performance of \textbf{ours} to that of \textbf{Vanilla}.}
        \label{tab:e2e_latency}
        \renewcommand{\arraystretch}{1.15}
        \setlength{\tabcolsep}{5pt}
        \resizebox{\linewidth}{!}{
        \begin{tabular}{l c c c c}
            \toprule
            \textbf{Benchmark}
              & \makecell{\textbf{Ours}\\[-2pt]{\small ($B_{\mathrm{token}}\!=\!256$)}}
              & \textbf{Vanilla}
              & \textbf{Speedup}
              & \makecell{\textbf{Rel.\ Perf.}\\[-2pt]{\small (Acc@1)}} \\
            \midrule
            HRBench8K
              & 2.21\,s & 11.26\,s
              & $5.10\times$
              & \textcolor{DropGreen}{96.3\%\;$\downarrow$} \\
            V$^{*}$
              & 1.85\,s & 5.08\,s
              & $2.75\times$
              & \textcolor{DropGreen}{93.4\%\;$\downarrow$} \\
            TreeBench
              & 1.97\,s & 5.38\,s
              & $2.74\times$
              & \textcolor{GainRed}{100.1\%\;$\uparrow$} \\
            MRWL
              & 1.97\,s & 5.90\,s
              & $3.00\times$
              & \textcolor{GainRed}{110.2\%\;$\uparrow$} \\
            \bottomrule
        \end{tabular}
        }
    \end{minipage}%
    \hfill
    %% ---- Right: Bar chart ----
    \begin{minipage}[t]{0.38\textwidth}
        \vspace{0pt}
        \centering
        \includegraphics[width=0.80\linewidth]{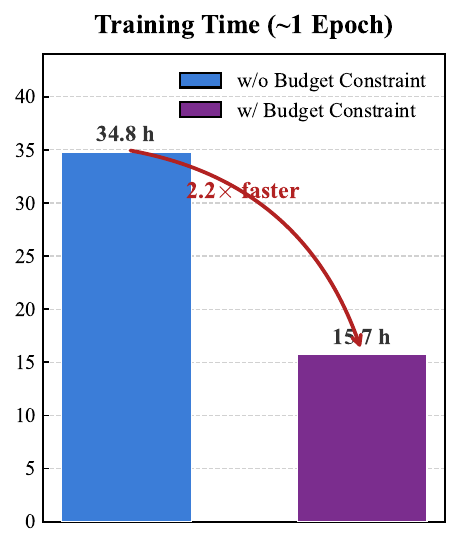}
        \captionof{figure}{RL training cost comparison.}
        \label{fig:training_cost}
    \end{minipage}
\end{figure*}
Beyond task performance, our budget-constrained paradigm yields substantial efficiency gains at both inference and training time.
\paragraph{\textbf{Inference efficiency.}} Table~\ref{tab:e2e_latency} compares the end-to-end inference latency per sample between our budget-constrained agent ($B_{\mathrm{token}}\!=\!256$) and the vanilla unconstrained baseline, measured under identical evaluation infrastructure.
By dynamically allocating visual tokens within a fixed budget, our method achieves \textbf{2.7--5.1$\times$ speedup} across all benchmarks, with the largest gain on HRBench8K where the high-resolution input incurs the heaviest visual token overhead for the unconstrained agent.
Notably, this acceleration comes at \emph{minimal} accuracy trade-off: on two of four benchmarks (TreeBench and MRWL), the constrained agent even \emph{surpasses} the baseline, suggesting that the budget constraint acts as an implicit regularizer that discourages redundant visual exploration.

\paragraph{\textbf{Training efficiency.}} The efficiency advantage also manifests during reinforcement learning. Fig.~\ref{fig:training_cost} plots the total training time cost one epoch of RL training. Compared to our budget-constrained RL, training data under no-constrain for one epoch costs $2.2\times$ more time with the same training configuration(19k training data of high resolution images, 6 A800 GPUs), primarily due to the reduced number of visual tokens processed. These empirical results highlight the potential of a budget-aware approach and may inspire both the research and industrial communities to consider adopting a new multimodal post-training paradigm.

\section{Related Work}
\subsection{Vision-Language Models}
Modern vision-language understanding builds on a representation foundation laid by contrastive learning at web scale. CLIP~\cite{radford2021learning} and ALIGN~\cite{jia2021scaling} show that aligning visual and textual embeddings through large-scale image-text pairing produces highly generalizable features, motivating a subsequent line of work that couples frozen large language models (LLMs) with pre-trained vision encoders via lightweight connectors~\cite{alayrac2022flamingo, li2023blip, liu2023visual}. A pivotal catalyst in this progression is visual instruction tuning introduced by LLaVA~\cite{liu2023visual}, which reframes multimodal learning as an instruction-following problem and enables LLMs to process interleaved image-text inputs. Both open-source community~\cite{bai2023qwen,zhu2025internvl3} and proprietary systems~\cite{geminiteam2025geminifamilyhighlycapable} scaled and refined this paradigm by incorporating extensive supervised fine-tuning with reinforcement learning from human or automated feedback. Despite the progress, modern VLMs are prone to hallucinations when they are asked for vague and small visual details~\cite{li2023evaluating, li2025analyzing}, and tend to rely disproportionately on textual priors rather than genuinely grounding answers in image content~\cite{zhang2025modalities, agrawal2025towards, liu2025more}. This gap between strong language-side reasoning and brittle visual perception forms a central motivation for our work.

\subsection{Tool Augmented Visual Reasoning} 
To address visual hallucination and over-reliance on textual priors, recent research has explored equipping VLMs with external tools for image manipulation, advancing visual reasoning from static image-text understanding to dynamic interactive interaction. Early works adopted training-free approaches: SEAL~\cite{wu2023vguidedvisualsearch} uses guided visual search directed by LLMs to focus on task-relevant regions, while ZoomEye~\cite{shen-etal-2025-zoomeye} employs a tree-search algorithm to mimic human-like zooming behavior. More recently, learning-based methods have emerged, training models to integrate tool manipulation into their perceptual processes. For instance, DeepEyes~\cite{deepeyes} employs end-to-end reinforcement learning to acquire tool-use behaviors, while other works~\cite{pixelreasoner,zhang2025adaptivechainoffocusreasoningdynamic} adopt a commonly used SFT-RL two-stage training pipeline. Despite these advances, recent studies~\cite{virl, pixelreasoner} reveal that tool-augmented models still suffer from \emph{lazy perception} -- relying on textual
shortcuts rather than genuinely leveraging visual tools for reasoning. To this end, we propose a new training paradigm that incentivizes genuine active perception through a constrained visual environment.
\section{Conclusion}
We presented \textbf{Starve to Perceive}, a training paradigm that enforces constrained visual bandwidth throughout both supervised fine-tuning and reinforcement learning, transforming active perception from an optional strategy into a necessary survival mechanism. Our experiments demonstrate that perceptual skills forged under token starvation transfer robustly to unconstrained evaluation—achieving the highest overall score (64.59) among all 7B tool-augmented baselines—while degrading only 0.18 points under strict budgets, compared to near 7-point drops for baselines trained without visual constraints. The paradigm simultaneously yields over 50\% reduction in RL training time and 2.7–5.1× inference speedup, requires no architectural changes or auxiliary losses, and integrates into standard post-training pipelines as a plug-and-play component. These findings suggest that superior visual reasoning emerges not from input density but from the optimization pressure to extract information efficiently, and we hope this perspective encourages the community to treat visual bandwidth as a deliberate design variable for building perceptually grounded vision-language agents.

% \clearpage\mbox{}Page \thepage\ of the manuscript.
% \clearpage\mbox{}Page \thepage\ of the manuscript.
% \clearpage\mbox{}Page \thepage\ of the manuscript.
% \clearpage\mbox{}Page \thepage\ of the manuscript.
% \clearpage\mbox{}Page \thepage\ of the manuscript. This is the last page.
% Now we have reached the maximum length of an ECCV \ECCVyear{} submission (excluding references and acknowledgements).
% References should start immediately after the main text, but can continue past p.\ 14 if needed. 
% \clearpage  % TODO FINAL: This \clearpage needs to be removed from both review and camera-ready versions.

% \section*{Acknowledgements}
% Please insert your acknowledgments here.

\bibliographystyle{splncs04}
\bibliography{main}

% ========== SUPPLEMENTARY MATERIAL STARTS HERE ==========
\clearpage 
\appendix  

\section*{Supplementary Material}
% Supplementary: Limitations Section
\section{Limitations}
\label{sec:supp_limitations}
While \textbf{Starve to Perceive} achieves strong gains with minimal modifications to existing post-training pipelines, three limitations merit acknowledgment. First, current evaluation concentrates on high-resolution visual search, chart and table understanding, and perception-intensive reasoning; broader validation across domains such as mathematical reasoning, OCR parsing, hallucination benchmarks, medical imaging, and satellite remote sensing remains as future work. Second, we have not yet evaluated the scalability of our budget-constrained inference pipeline under extended multi-turn reasoning regimes in both training and evaluation; future work could explore adaptive turn budgets that terminate early when model confidence is high and extend when uncertainty persists. Third, the resolution-conditioned token budget $B(X)$ is governed by a fixed compression rate $\gamma$ and static bounds $B_{\min}$, $B_{\max}$; an online curriculum that tightens $\gamma$ as the model's success rate rises could maintain the learning signal at the capability frontier and improve sample efficiency. 

% Supplementary: Data and Training Details
\section{Data and Training Details}
\label{sec:supp_data_training}

\subsection{Training Details}
For Budget-Aware Visual Instruction Tuning, we use LLaMA-Factory~\cite{zheng2024llamafactory} on $4\times$ A800 (80\,GB) GPUs for 3 epochs with default LoRA fine-tuning settings and a batch size of 32. For reinforcement learning, we use VeRL~\cite{sheng2024hybridflow} with 6 GPUs for policy training and 2 GPUs for the LLM-as-judge reward model, employ near on-policy updates where the behavior policy is synchronized with the improvement policy after each training step, set the training batch size to 216 query-response pairs with a PPO mini-batch size of 36, use a constant learning rate of $1\times 10^{-6}$, and set the maximum interaction turns during rollout to 5. We set the resolution-conditioned token budget parameters to $\gamma=6.25$, $B_{\min}=169$, and $B_{\max}=1337$, where $B_{\max}$ approximately matches the visual token consumption of a 1024$\times$1024 image with patch size 28 to keep high-resolution inputs within the constrained observation space, and $\gamma=6.25$ enforces a consistent degree of perceptual starvation across resolutions while preserving a viable learning signal. We implement a partial version of DAPO~\cite{dapo} on top of GRPO, set $c_{\mathrm{clip\_high}}=0.28$ to encourage exploration, and use dynamic sampling to improve training efficiency and stability under sparse reward signals.

\subsection{RL Training Data}
We apply two filtering steps before training to ensure the visual bandwidth constraint creates meaningful optimization pressure. First, we require images to have resolution greater than $512 \times 512$, because large images widen the gap between the constrained low-resolution view and the original high-resolution image, placing greater demand on precise localization and making the passive shortcut strategy less viable. Second, for each question, we sample $n=8$ responses from the base model and compute the mean accuracy across these responses. We retain questions with accuracy $\in [0.125, 0.375]$, because queries that are too easy (accuracy $> 0.375$) yield saturated reward signals that provide little gradient information, while queries that are too hard (accuracy $< 0.125$) generate mostly zero-reward trajectories that fail to distinguish between effective and ineffective perception strategies. The intermediate difficulty range ensures that the RL signal remains informative and that the model operates near its capability boundary where the learning signal is strongest.

\subsection{SFT Data Construction}
\label{subsec:sft_construction}
We curate the SFT training data via two approaches:teacher distillation and  trajectory rewriting. For teacher distillation, we use rejection sampling to generate and select high-quality trajectories from the base model. For trajectory rewriting, we reuse the SFT datasets from two existing works (TreeVGR~\cite{treebench} and VisualProbe~\cite{minio3}). Each of them requires different pipelines due to their different source formats. 

\paragraph{TreeVGR: Passive Grounding $\to$ Active Multi-Turn.}
TreeVGR-SFT-37K trajectories are single-turn passive grounding responses of the form:
\begin{lstlisting}[language={},breaklines=true,basicstyle=\ttfamily\footnotesize,columns=fullflexible,keepspaces=true]
<think>...<box>[x1,y1,x2,y2]</box>...</think><answer>...</answer>
\end{lstlisting}
where bounding boxes are absolute-pixel annotations embedded in the chain-of-thought---not tool calls.
We apply two filters before rewriting.
\textbf{(1) Self-correction filter:} we retain only samples that contain the self-correction signal, as these provide explicit training signal for recovering from erroneous zoom attempts and are the primary target for the structural rewriting.
\textbf{(2) Second-order grounding filter:} we discard samples where a bounding box is grounded relative to a previously cropped region rather than the original image, which would introduce ambiguous coordinate frames that cannot be faithfully preserved during rewriting.

The filtered samples are then rewritten into active multi-turn \texttt{focus} trajectories by GPT-4o with few-shot prompting (see \cref{subsec:rewriting_prompt} for the exact prompt).
This is a \emph{structural} transformation requiring semantic understanding of the original reasoning (\eg, deciding how to sequence tool calls and how to narrate the self-correction step).

After rewriting, each output is validated before being retained.
The checks enforce: (1) at least 3 turns with strictly alternating \texttt{gpt}/\texttt{human} roles; (2) every GPT turn contains \texttt{<think>...</think>}; (3) the number of \texttt{<image>} tokens in each tool response matches the number of bboxes in the preceding tool call (1--3); (4) the multiset of all bboxes used across tool calls exactly matches the multiset extracted from the original \texttt{<box>} annotations.
Trajectories failing any check are discarded.

\paragraph{VisualProbe: Coordinate-Space Alignment.}
VisualProbe (Mini-o3) trajectories already use the active multi-turn \texttt{focus} format with absolute pixel coordinates referenced to the \emph{full-resolution} image.
Our training environment, however, constrains the model's view to a downscaled overview image ($B=256$). Since the agent observes in the overview image space, rewriting is therefore purely geometric: given scale factors $(s_x, s_y) = (W_\text{ov}/W_\text{orig},\,H_\text{ov}/H_\text{orig})$, every bbox $[x_1,y_1,x_2,y_2]$ is transformed to $[\lfloor x_1 s_x\rceil, \lfloor y_1 s_y\rceil, \lfloor x_2 s_x\rceil, \lfloor y_2 s_y\rceil]$ and all coordinate mentions in \texttt{<think>} text are updated by string matching.
The overview image and per-bbox cropped observations (each also constrained to $B_\text{pix}$) are pre-rendered and stored alongside the trajectory.

Before rescaling, we apply three filters to the raw trajectories:
\textbf{(1) Trajectory length:} discard samples with more than 3 zoom turns, as longer trajectories tend to contain redundant search steps.
\textbf{(2) Grounding order:} exclude second-order grounding samples where the tool call is grounded in a previously cropped observation rather than the overview image, which would introduce coordinate-frame ambiguity after rescaling.
\textbf{(3) Single-pass filter:} exclude trajectories with no tool call, ensuring every retained example exercises the active perception loop.
This yields 2,545 budget-aware trajectories.

\paragraph{Teacher Distillation: Budget-Aware Visual Instruction Tuning.}
Using seed VQA data, we perform rejection sampling under the same visual budget constraint to obtain high-quality budget-aware trajectories. For each question $Q$, we sample $k$ rollouts from a (stronger) teacher policy in the constrained environment and retain only trajectories whose final answer matches the ground-truth. We then filter redundant zoom steps by measuring the overlap between any two selected regions: for region bounding boxes $b_i$ and $b_j$, we compute $\mathrm{IoU}(b_i,b_j)=\frac{|b_i\cap b_j|}{|b_i\cup b_j|}$ and discard trajectories where two turns select highly overlapping regions (i.e., the model effectively focuses on the same area twice). Finally, we keep one remaining trajectory per question.

\subsection{Data Composition}

\paragraph{\textbf{SFT (Budget-Aware Visual Instruction Tuning).}}
Our SFT corpus of 7,152 budget-aware trajectories is assembled from three sources. \textbf{PixelReasoner}~\cite{pixelreasoner} (2,348 trajectories) contributes high-quality multi-turn visual search and document understanding trajectories; we extract the queries from the SFT dataset, then we use a stronger teacher with better instruction following capacities to generate new reasoning trajectories. After that, we exclude all single-pass (non-zooming) trajectories to ensure every training example exhibits genuine active perception behavior. \textbf{VisualProbe}~\cite{minio3} (2,545 trajectories) provides trajectories originally grounded in full-resolution images; each trajectory is converted to the budget-aware format via rule-based coordinate rescaling followed by LLM-based reasoning rewriting (GPT-4o) to align the thinking process with the compressed observation space. \textbf{TreeVGR}~\cite{treebench} (2,259 trajectories) is drawn from the TreeVGR-SFT-35K dataset, whose trajectories use relative coordinates in a passive single-turn grounding format; we convert them to our active multi-turn absolute-pixel format using GPT-4o prompted with the trajectory rewriting procedure described in \cref{sec:supp_prompts}. Since TreeVGR contains many low-resolution images that already fit within the token budget, no downscaling is applied for this subset.

\paragraph{\textbf{RL (Reinforcement Learning).}}
Our RL training pool of 19,021 queries is drawn from four sources covering diverse visual reasoning demands. \textbf{ViRL5k}~\cite{virl} (2,722 queries) spans general understanding, visual grounding, and OCR tasks, providing broad task coverage. \textbf{DeepEyes}~\cite{deepeyes} (5,000 queries, chart subset only) supplies chart and figure reasoning queries, which require precise localization of legends, axes, and data points that are naturally impossible to resolve from the compressed global view alone. \textbf{VisualProbe}~\cite{minio3} (5,729 queries) focuses on small-target detection and high-resolution visual search, directly aligned with our training objective. \textbf{TreeVGR}~\cite{treebench} (5,570 queries) contributes traceable visual grounding tasks with verified intermediate evidence, offering supervision signal for structured multi-step visual reasoning. All queries pass the resolution and difficulty filters described above before being included in training.

\section{Qualitative Analysis}
\label{sec:supp_qualitative}

\begin{figure}[t]
  \centering
  \IfFileExists{sup/Figs/qualitative.pdf}{%
    \includegraphics[width=\linewidth]{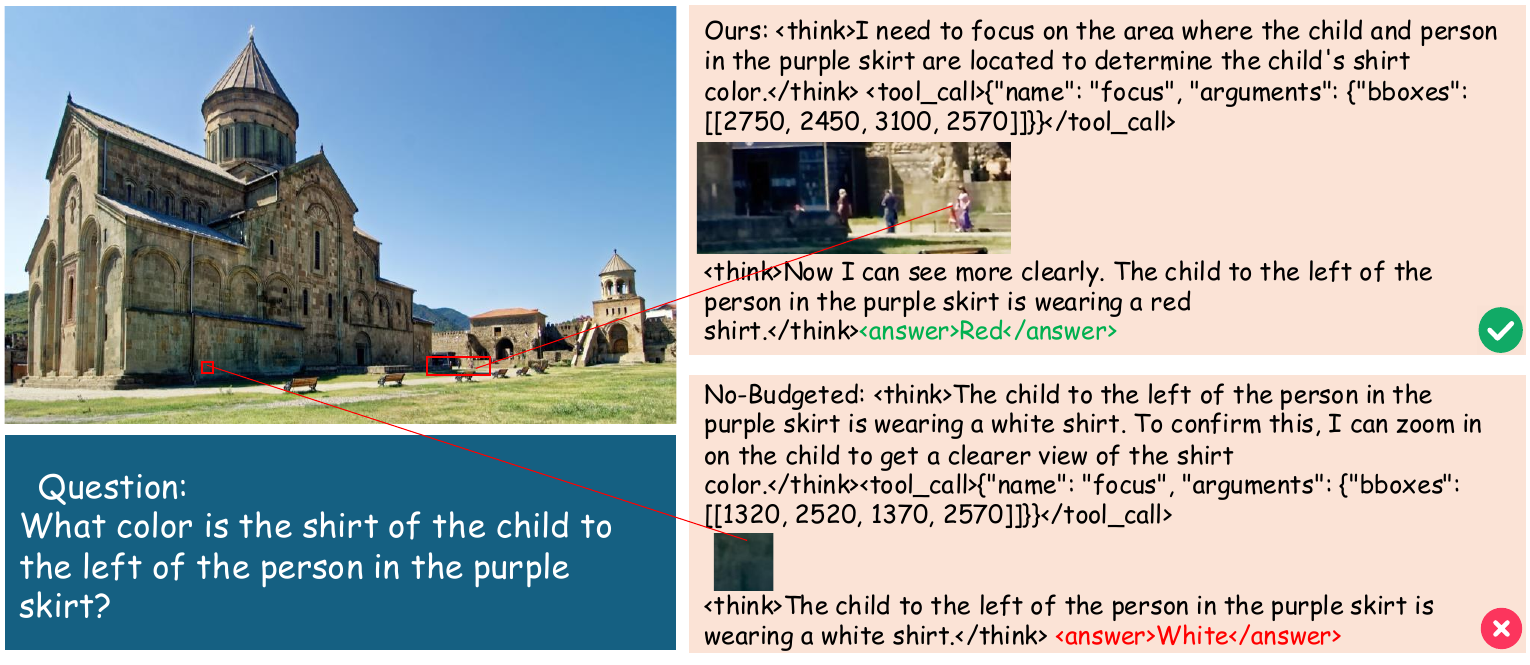}%
  }{%
    \includegraphics[width=\linewidth]{Figs/qualitative.pdf}%
  }
  \caption{Qualitative example of active perception. Our budget-aware model focuses on informative regions and grounds its answer in the returned local evidence, while the non-budgeted baseline exhibits a lazy-perception failure.}
  \label{fig:supp_qualitative}
\end{figure}

Figure~\ref{fig:supp_qualitative} shows a representative active-perception trajectory generated under the same interaction format used in our training and evaluation pipeline. The model first reasons over the budgeted overview image, decides whether high-resolution local evidence is needed, calls the \texttt{focus} tool with an explicit bounding box, and then produces the final answer after observing the returned crop. This example illustrates the kind of budget-aware behavior encouraged by our data construction and reinforcement learning pipeline.{\color{white}\cite{li2025chemvlm,li2025iag,zhang2025critic}}

In the shown case, the budgeted model selects a region covering both the queried child and the reference person in the purple skirt, allowing it to resolve the shirt color from local visual evidence. By contrast, the non-budgeted baseline focuses on a less informative region and preserves a premature answer, a typical ``lazy perception'' failure mode in which the model appears to use the tool but does not acquire the decisive evidence. This qualitative comparison complements the quantitative gains by showing how visual bandwidth constraints encourage more evidence-seeking region selection. {\color{white}\nocite{wang2025emergent,wang2025reverse,wang2026rationalrewards,wang2025vl,zhang2026instruction,zhang2505evaluating,wang2025code,wang2026badseeingbadthinking}}

\section{Budget Sweep Analysis}
\label{sec:supp_budget_sweep}

To examine whether the gains are tied to a single manually chosen visual budget, we evaluate our model and the strongest internal baseline, BA-SFT+VanillaRL, under multiple inference-time budgets $B \in \{128,256,512,1024\}$. For each budget, we average accuracy over the same 11 benchmarks used in the main evaluation and report both the average score and the number of benchmark splits where our model wins head-to-head.

\begin{table}[t]
  \centering
  \small
  \setlength{\tabcolsep}{4.5pt}
  \caption{Budget sweep comparing our budget-aware model with the strongest internal baseline. H2H denotes head-to-head wins across 11 benchmark splits.}
  \label{tab:supp_budget_sweep}
  \begin{tabular}{c c c c c}
    \toprule
    $B$ & Ours Avg & Vanilla Avg & $\Delta$ (Ours$-$Vanilla) & H2H Wins \\
    \midrule
    128  & 47.58 & 46.20 & +1.38 & 8/11 \\
    256  & 52.35 & 51.54 & +0.81 & 7/11 \\
    512  & 56.82 & 56.32 & +0.50 & 8/11 \\
    1024 & 60.69 & 60.09 & +0.60 & 7/11 \\
    \bottomrule
  \end{tabular}
\end{table}

Table~\ref{tab:supp_budget_sweep} shows that the budget-aware model maintains consistent advantages across the tested budget range. The average gain is largest under the strictest budget ($B=128$), where visual starvation imposes the strongest pressure to select informative regions, but the advantage remains positive even as the budget increases to $1024$. This trend suggests that the observed gains are not an artifact of a single budget setting; rather, training under constrained visual bandwidth improves robustness across a range of inference-time starvation levels.

% Supplementary: Prompts Section
\section{Prompts}
\label{sec:supp_prompts}

\subsection{Prompt for RL Training}
\label{subsec:rl_prompt}

The system prompt equips the model with the \texttt{focus} tool:

\begin{lstlisting}[language={},caption={RL Training System Prompt},breaklines=true,basicstyle=\ttfamily\footnotesize,columns=fullflexible,keepspaces=true]
You are a helpful assistant.

# Tools
You are provided with the function signature within <tools></tools> XML tags:
<tools>
{
  "type": "function",
  "function": {
    "name": "focus",
    "description": "Request a high-resolution local region of the first image and zoom in",
    "parameters": {
      "type": "object",
      "properties": {
        "bboxes": {
          "type": "array",
          "minItems": 1,
          "maxItems": 3,
          "items": {
            "type": "array",
            "items": { "type": "integer" },
            "minItems": 4,
            "maxItems": 4,
            "description": "The bounding box of the region to crop, as [x1, y1, x2, y2] in ABSOLUTE PIXEL COORDINATES of the first image."
          },
          "description": "A list of bounding boxes to zoom in on. You can request 1-3 bboxes at a turn."
        }
      },
      "required": ["bboxes"]
    }
  }
}
</tools>

# How to call a tool
Return a json object with function name and arguments within <tool_call></tool_call> XML tags:
<tool_call>
{"name": <function-name>, "arguments": <args-json-object>}
</tool_call>

Example:
<tool_call>
{"name": "focus", "arguments": {"bboxes": [[10, 20, 100, 200]]}}
</tool_call>
\end{lstlisting}

The user prompt instructs the model to reason before acting and invoke \texttt{focus} only when needed:

\begin{lstlisting}[language={},caption={RL Training User Prompt},breaklines=true,basicstyle=\ttfamily\footnotesize,columns=fullflexible,keepspaces=true]
Think first, call focus if needed, then answer if you are confident.
Format strictly as:
  <think>...</think>
  <tool_call>...</tool_call>  (if tools needed)
  <answer>...</answer>
You should continue your reasoning process within <think>...</think>
based on the content returned by the function tool.
Here is the question:
\end{lstlisting}

\subsection{Prompts for Trajectory Rewriting}
\label{subsec:rewriting_prompt}

The two SFT subsets use different rewriting procedures; see \cref{subsec:sft_construction} for the full pipeline and selection criteria.

\paragraph{TreeVGR Rewriting Prompt.}
GPT-4o is prompted with the following system prompt (plus three in-context examples) to convert passive single-turn grounding responses into active multi-turn \texttt{focus} trajectories:

\begin{lstlisting}[language={},caption={TreeVGR Rewriting System Prompt},breaklines=true,basicstyle=\ttfamily\footnotesize,columns=fullflexible,keepspaces=true]
You are an expert data synthesist for training visual agents.
Your task is to convert "Passive Grounding" data (descriptions
with bounding boxes) into "Active Visual Search" training data
AS IF the agent is equipped with the focus tool.

Input: A User Question and a GPT Response of the form
  <think>...<box>[x1,y1,x2,y2]</box>...</think><answer>...</answer>

Output Goal: Rewrite the response into a multi-step format where:
  1. The agent thinks about needing to examine the region closely.
  2. The agent calls focus on the specific region(s) from the input.
  3. If a box is marked "Wait, this box seems wrong.", the agent
     self-corrects by pivoting to other regions.

Strict Rules:
- PRESERVE all bounding box coordinates exactly.
- ALL <box>...</box> bboxes must appear in <tool_call>...</tool_call>.
- Each turn: <think>...</think> <tool_call>...</tool_call>
         OR: <think>...</think> <answer>...</answer>
- Bboxes per tool_call: 1-3. Must match <image> count in tool_response.
- <box> and </box> must NEVER appear in the output.

Output Format: strict ShareGPT multi-turn JSON (3-5 turns):
[
  {"from":"gpt",   "value":"<think>...</think> <tool_call>...</tool_call>"},
  {"from":"human", "value":"<tool_response><image>...</tool_response>"},
  {"from":"gpt",   "value":"<think>...</think><answer>...</answer>"}
]
\end{lstlisting}

\paragraph{VisualProbe Rewriting.}
VisualProbe trajectories are already in active \texttt{focus} format; rewriting is a purely rule-based coordinate rescaling from full-resolution to overview pixel space. No LLM prompt is involved; see \cref{subsec:sft_construction} for details.

\subsection{Teacher Distillation Prompts}
We use a more detailed prompt version comparing to the RL training prompt for generating teacher trajectories under budget constrain visual inference environments. The prompt is as follows:

\begin{lstlisting}[language={},caption={Teacher Distillation Prompt},breaklines=true,basicstyle=\ttfamily\footnotesize,columns=fullflexible,keepspaces=true]
You are a helpful assistant.
A user gives a image with a question. Your task is to solve the question based on the **Fixed Retina** constraint:
- MAX_VIEW_PIXELS = 28 * 28 * 16 * 16 pixels.
- The user's image is compressed into an `overview image` with a maximum resolution of MAX_VIEW_PIXELS.
- You can call the **focus** tool to request detailed views for specific regions. Both overview and focused regions are constrained by MAX_VIEW_PIXELS.
You should perform the `focus` search until you are completely SURE that the question can be solved. 

# Tools
You are provided with the function signature within <tools></tools> XML tags:
<tools>
{
    "type": "function",
    "function": {
        "name": "focus",
        "description": "Request a detailed view of the overview image from the original image pixel space. The returned focused image will still be contrained to MAX_VIEW_PIXELS if too large.",
        "parameters": {
            "type": "object",
            "properties": {
                "bboxes": {
                    "type": "array",
                    "minItems": 1,
                    "maxItems": 3,
                    "items": {
                        "type": "array",
                        "items": {
                            "type": "integer"
                        },
                        "minItems": 4,
                        "maxItems": 4,
                        "description": "The bounding box of the region to crop, as [x1, y1, x2, y2] in ABSOLUTE PIXEL COORDINATES of the overview image."
                    },
                    "description": "A list of bounding boxes to zoom in on. You can request 1-3 bboxes at a turn."
                }
            },
            "required": ["bboxes"]
        }
    }
}
</tools>
# How to call a tool
Return a json object with function name and arguments within <tool_call></tool_call> XML tags:
<tool_call>
{"name": <function-name>, "arguments": <args-json-object>}
</tool_call>

**Example**:  
<tool_call>  
{"name": "focus", "arguments": {"bboxes": [[10, 20, 100, 200]]}}  
</tool_call>
\end{lstlisting}
% Supplementary bibliography (if needed separately)
% Or skip if you want shared bibliography

\end{document}